# The Most Persistent Soft-Clique in a Set of Sampled Graphs

**Novi Quadrianto**                                    NOVI.QUADRIANTO@GMAIL.COM
University of Cambridge, Cambridge, UK

**Chao Chen**                                          CHAO.CHEN@IST.AC.AT
IST Austria (Institute of Science and Technology Austria), Klosterneuburg, Austria

**Christoph H. Lampert**                               CHL@IST.AC.AT
IST Austria (Institute of Science and Technology Austria), Klosterneuburg, Austria

## Abstract

When searching for characteristic subpatterns in potentially noisy graph data, it appears self-evident that having multiple observations would be better than having just one. However, it turns out that the *inconsistencies* introduced when different graph instances have different edge sets pose a serious challenge. In this work we address this challenge for the problem of finding *maximum weighted cliques*.

We introduce the concept of *most persistent soft-clique*. This is subset of vertices, that 1) is almost fully or at least densely connected, 2) occurs in all or almost all graph instances, and 3) has the maximum weight. We present a measure of clique-ness, that essentially counts the number of edge missing to make a subset of vertices into a clique. With this measure, we show that the problem of finding the most persistent soft-clique problem can be cast either as: a) a max-min two person game optimization problem, or b) a min-min soft margin optimization problem. Both formulations lead to the same solution when using a partial Lagrangian method to solve the optimization problems. By experiments on synthetic data and on real social network data we show that the proposed method is able to reliably find soft cliques in graph data, even if that is distorted by random noise or unreliable observations.

## 1. Introduction

Graphs are used ubiquitously in computer science in order to represent data objects and their interrelations. Consequently, machine learning and data mining research has developed a large number of methods to analyze given graph structures and to identify substructures of predefined properties, in particular *cliques*, i.e. subsets of vertices that are fully connected with respect to the graph's edge set.

In this work, we extend this reasoning to the case where multiple, potentially noisy or incomplete, instances of a graph are available for analysis. The hard criterion of a set of vertices being fully connected becomes too limiting in this case, so instead we look for *persistent soft-cliques*, i.e. subgraphs that are *almost fully* or at least *densely* connected, and that persists through all or most instances of the graph. For given several instances of a *weighted* graph, we are interested to find a persistent soft-clique with the highest weight. We call this a *most persistent soft-clique* problem.

By solving the most persistent clique problem we can extend a wide range of applications that relied on finding cliques in graphs to situations where a collection of graphs is available, e.g. measurements at different points of times, but where each graph instance might have a different edge set, e.g. due to noisy or incomplete observations. These noisy snapshots of the same graph pose challenging tasks related to inconsistent patterns, but could give us more confidence in characterizing the inherent pattern or phenomenon. Take as an illustrative example, the usage of dense subgraphs in mobile phone or location-based social networks to identify groups of friends or families. A temporal dimension arises naturally for such a graph where, for example, different hours in a day lead to several samples of the graphs. It is reasonable to assume that dense subgraphs that appear in all of the samples of the graphs are the groups of friends or families that we would like to identify. However, it can also be expected that in each individual observation of the graphs not every person will be observed within the subgraph: he or she could have left the group temporarily due to other commitments, or the measurement itself could be faulty, e.g. due to a network outage.

**Contribution** In this work, we make the following contributions: 1) We introduce a simple clique-ness measure that is suitable for finding a persistent soft-clique across multiple noisy instances of a graph. Intuitively, the measure counts and penalizes the number of edges missing to make the selected subset of vertices into a clique. 2) We show how this measure can be used in an optimization framework either as a two person game or as a relaxation of the hard-clique problem with slack variables, to find a maximum weighted persistent soft-clique. 3) We show that both formulations lead to the same solution when using a partial Lagrangian method to solve the optimization problems: the upper bound obtained for the max-min two player game formulation coincides with the lower bound of the min-min slack formulation. 4) We perform an experiment on synthetic data and provide an application in social network data where the graphs are sampled across time.

First, however, we give a short overview of related work to provide some context for our contributions.

## 1.1. Related Work

The first work on the use of cliques in social network graphs by far predates recent efforts in machine learning, data mining or network science. Luce & Perry (1949) studied adjacency matrices of *friendship* graphs in order to identify *cliques* of friends. They already observed that the criterion of all vertices being connected to each other can be overly strict, and they introduced the softer *n-clique* criterion. Later, Alba (1973) showed, however, some non-intuitive facts about *n*-cliques, e.g. that their elements might be completely disconnected with respect to the original graph structure. Afterwards, many alternative constructions were introduced, see, e.g., (Scott, 1988) for a textbook overview, and (Brunato et al., 2008; Robardet, 2009) for some recent developments. Besides the social sciences, other areas of research that deal with network structured data have adopted the search for cliques as part of their research methodology, in particular *bioinformatics*, *chemistry* and *(electrical) engineering*. *Mathematics*, on the other hand, studied properties of graphs and subgraphs more formally, through the anal-ysis of function optima defined on the graph (Motzkin & Straus, 1965), or statistical properties of graphs with random edge sets (Bollobás, 2001). In *theoretical computer science* and *operations research* graph-based algorithms are today objects of core interest, e.g. in the analysis of their computational complexity, or as tools to abstractly model the process of computation in a computer itself.

With the rise of *machine learning* and *data mining*, a lot of more applied work on identifying structure within graphs of empirical data has emerged, see, e.g., (Schaeffer, 2007) for a survey. For example, Gupta & Ghosh (2005) and Crammer et al. (2008) looked at a one-class clustering or classification problem which, given a dataset, aims at identifying a coherent yet small subset of data points. This one-class clustering problem finds itself applications in bioinformatics to find gene modules and natural language to extract documents' topics. Pavan & Pelillo (2007) use the weighted version of maximal (non-extendable) clique to perform a pairwise clustering. In a recent time, the maximum weighted clique concept has also found increased use in *computer vision* problems for performing image segmentation tasks, for example (Brendel & Todorovic, 2010; Ion et al., 2011).

While all the above learning methods rely on only a single (noisy) graph to be analyzed, it is intuitive that having access to multiple observations should make the result of the analysis more robust against random effects, missing observations, and outliers. Achieving this intuition is our goal for the rest of the paper.

## 2. Persistent Soft-Cliques

In this section we formulate our main contributions: a measure how close a set of vertices is to being a clique, and its generalization to multiple graph instances. First, we introduce the necessary notation for this. In Section 2.1, we cast the problem of finding a maximal clique in a single graph as an integer optimization problem, and in Section 2.2 we extend this notion to the case of soft-cliques persisting across multiple instances of a graph.

**Notation** For a set of vertices $\mathcal{V} = \{v_1, \ldots, v_n\}$, let $\mathcal{E}_t \subseteq \mathcal{V} \times \mathcal{V}$, for $t = 1, \ldots, T$, be multiple observed sets of edges, and let $k_t : \mathcal{V} \times \mathcal{V} \to \mathbb{R}_+$ denote non-negative weight functions between the vertices of $\mathcal{V}$. We form weighted graphs $\mathcal{G}_t = (\mathcal{V}, \mathcal{E}_t, k_t)$, for $t = 1, \ldots, T$. In the following, we will refer to $t$ as a 'time' variable, but note that this is meant in a broad sense: while $t = 1, \ldots, T$ can correspond to an actual temporal sequence, it can also just denote different instances of a graphs in arbitrary order. Furthermore, we will interpret the values of $k_t(v_i, v_j)$ as measures of similarity between vertices, e.g. given by a positive definite kernel function (Schölkopf & Smola, 2001), but the method we describe does not depend on $k_t$ having any specific structure.

## 2.1. Soft Clique-ness

In the following, we will define a measure of *cliqueness*. Let $S \subset \mathcal{V}$ be a vertex subset. For this, we identify $S$ with its binary indicator vector $x_i \in \{0,1\}^{|\mathcal{V}|}$, where $x_i = 1$ if $v_i \in S$, and $x_i = 0$ otherwise. The problem of finding a *maximum weighted clique* in a weighted graph $(\mathcal{V}, \mathcal{E}, k)$ can then be cast as the following integer optimization problem (Pardalos & Xue, 1994):

$$\max_{x \in \{0,1\}^{|\mathcal{V}|}} \sum_{1 \le i < j \le n} x_i x_j k(v_i, v_j) \tag{1a}$$

$$\text{subject to} \sum_{1 \le i < j \le n} x_i x_j \mathbb{I}[(i,j) \notin \mathcal{E}] = 0. \tag{1b}$$

In the above, we make use of Iverson's bracket notation: $\mathbb{I}[P] = 1$ for the condition $P$ is true and it is 0 otherwise. The constraint enforces that any two variables in the selected subset are connected by an edge, i.e. the subgraph given by $x$ is a clique. For a non-negative weight function $k(\cdot, \cdot)$, the above optimization also ensures that the inferred clique is maximal, i.e. no extension of the clique by adding one or more vertices is possible.

For graphs based on noisy observations it is clear that the constraint (1b) is too strict: a single missing edge will exclude a subset of vertices from further consideration. Instead it makes sense to look for *soft-cliques*, i.e. subgraphs that are *almost fully* or at least *densely* connected. To formalize this, we observe that the left hand side of the clique-constraint (1b) has a simple interpretation: for any fixed $x \in \{0,1\}^{|\mathcal{V}|}$,

$$\beta := \sum_{1 \le i < j \le n} x_i x_j \mathbb{I}[(i,j) \notin \mathcal{E}] \tag{2}$$

counts how many edges are missing in $\mathcal{E}$ for the selected vertices to be a clique. Consequently, we will use Equation (2) as a measure of how far a set of vertices is from being a clique: for a set with $k$ elements, $\beta$ can take values between 0 ($x$ is a clique), and $\binom{k}{2}$ ($x$ is a completely disconnected), see Figure 1 for an illustration.

## 2.2. Persistency of a Clique over Time

Given multiple instances of a graph, $\mathcal{G}_t = (\mathcal{V}, \mathcal{E}_t, k_t)$, for $t = 1, \ldots, T$, finding a clique $x \in \{0,1\}^{|\mathcal{V}|}$ that persists through time is a straight-forward extension of the above single-graph case. We enforce the clique constraint (1b) simultaneously in all time slices, yielding the optimization problem

$$\max_{x \in \{0,1\}^{|\mathcal{V}|}} \sum_{1 \le i < j \le n} x_i x_j k(v_i, v_j) \tag{3a}$$

$$\text{subject to} \sum_{1 \le i < j \le n} x_i x_j \mathbb{I}[(i,j) \notin \mathcal{E}_t] = 0, \quad \forall t = 1, \ldots, T, \tag{3b}$$

where $k(v_i, v_j) = \sum_t k_t(v_i, v_j)$ encodes the total similarity of vertices during the time of consideration. Deriving an analogue soft-clique formulation is less straight-forward, since at different time steps, a different subset of edges might be missing to make a set of vertices a clique. In the next sections we present two ways to formalize the concept of a finding a soft-clique that persists over time. Subsequently, we will show that both formulation lead to identical solutions if solved in a Lagrangian relaxation framework.

**Slack Perspective** We start by making each of the hard-cliques constraints Equation (3b) into a soft-clique constraints by introducing *slack variables*, $\beta_t$, for $t = 1, \ldots, T$, on the right hand side. To avoid degenerate solutions, we penalize the resulting slack vector $\beta = (\beta_1, \ldots, \beta_T)$ in the objective function by a multiple of its $L^p$-norm for some $p \ge 1$. For reasons that will become clear later we furthermore replace the maximization by a minimization of the negative objective. This leads to the following optimization problem

$$\min_{x \in \{0,1\}^{|\mathcal{V}|}} \min_{\beta \in \mathbb{R}_+^T} -\sum_{1 \le i < j \le n} x_i x_j k(v_i, v_j) + \eta \|\beta\|_{L_p}^p \tag{4a}$$

$$\text{subject to} \sum_{1 \le i < j \le n} x_i x_j \mathbb{I}[(i,j) \notin \mathcal{E}_t] \le \beta_t \quad \forall t = 1, \ldots, T. \tag{4b}$$

Note that the form of (4) also allows interpretation as a *regularized risk functional*: $\|\beta_t\|_{L^p}^p$ is a loss term that measures how bad the choice of $x$ is as candidate for a most persistent soft-clique. The term $-\sum_{1 \le i < j \le n} x_i x_j k(v_i, v_j)$ acts as a regularizer that encourages the opposite direction: minimizing it requires as many vertices as possible to be selected, in particular those that have high similarity values. The variable $\eta$ is a trade-off parameter that controls the relative influence of loss and regularization terms. Intuitively, by solving (4) we look for the optimal balance between the goal of collecting as many similar vertices, and the goal to not include vertices that are frequently disconnected.

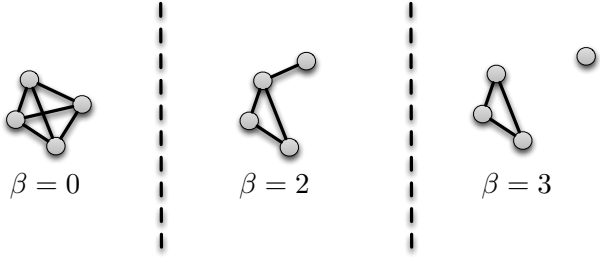

*Figure 1.* A soft measure of clique-ness. The measure counts the number of missing edges that will otherwise make the group of vertices a clique. The measure value is given assuming the size of the clique is 4.

**Two-Person Game Perspective** Alternatively, we can view the spatio-temporal cliques identification as a *game* of two competing players: *inlier* and *outlier*. The *inlier* player controls $x \in \{0,1\}^{|\mathcal{V}|}$ and aims at finding a group of variables with as large weight as possible. The *outlier* aims at reducing the objective value by controlling variables $\beta_1, \ldots, \beta_T$, which he or she can increase up a limit given by the number of edges missing to make $x$ a clique. Mathematically, this game is expressed by the following optimization problem:

$$\max_{x \in \{0,1\}^{|\mathcal{V}|}} \min_{\beta \in \mathbb{R}_+^T} \sum_{1 \le i < j \le n} x_i x_j k(v_i, v_j) - \sum_t \beta_t^p \quad (5a)$$

subject to, for all $t = 1, \ldots, T$,

$$\sum_{1 \le i < j \le n} x_i x_j \mathbb{I}[(i,j) \notin \mathcal{E}_t] \ge \beta_t \quad (5b)$$

for some $p \ge 1$. The intuition behind this resembles the one above: one player aims at forming sets with many similar vertices, the other player aims at excluding vertices if they are disconnected in many of the graph instances.

## 3. The Optimization

The key idea to solve the optimization problems in (4) and (5) is to relax the soft clique-ness constraint further. We replace the clique-ness constraint by a Lagrangian. This does not ensure that we will be able to meet the clique-ness constraints exactly anymore. We will instead only be able to state ex-post that the relaxed solution is optimal for the observed clique-ness distribution.

### 3.1. Lagrange Relaxations

In the following, we will show that slack and two-person game formulations lead to the same solution when using a partial Lagrangian method to solve the optimization problems: the *lower bound* of the min-min slack formulation coincides with the *upper bound* obtained for the two player game formulation.

**Lower Bound of Slack Perspective** By weak duality theorem (Boyd & Vandenberghe, 2004), we arrive at the following lower bound of (4):

$$\max_{\lambda \in \mathbb{R}_+^T} \min_{x \in \{0,1\}^{|\mathcal{V}|}, \beta \in \mathbb{R}_+^T} - \sum_{1 \le i < j \le n} x_i x_j k(v_i, v_j)$$

$$+ \eta \|\beta\|_{L_p}^p + \sum_t \lambda_t \left\{ \sum_{1 \le i < j \le n} x_i x_j \mathbb{I}[(i,j) \notin \mathcal{E}_t] - \beta_t \right\}. \quad (6)$$

**Upper Bound of Game Perspective** Similarly with the bound on slack perspective, we have the following upper bound of (5):

$$\min_{\lambda \in \mathbb{R}_+^T} \max_{x \in \{0,1\}^{|\mathcal{V}|}} \min_{\beta \in \mathbb{R}_+^T} \sum_{1 \le i < j \le n} x_i x_j k(v_i, v_j)$$

$$- \sum_t \beta_t^p + \sum_t \lambda_t \left\{ \sum_{1 \le i < j \le n} x_i x_j \mathbb{I}[(i,j) \notin \mathcal{E}_t] - \beta_t \right\} \quad (7)$$

A standard approach to dualize the problem of (6) and (7) is to eliminate the primal variables $x$ and $\beta$. We take the approach to partially dualize the problem by only finding the stationary point with respect to the primal variables $\beta$. In this paper, we give explicit formulation for the case of $p = 1$ and $p = 2$[1]. The $p = 1$ case leads to a maximum weighted clique in a single graph with modified weight functions, and the $p = 2$ case leads to a *series* of maximum weighted clique problems.

**$\ell_1$ Soft Clique-ness Measure** For simplicity of presentation, in what follows, we show the case for the trade-off parameter $\eta = 1$. In (6), taking the point where the gradient with respect to $\beta_t$ vanishes leads to the dual variables take the form of $\lambda_t = +1$ for $t = 1, \ldots, T$. Similarly, for (7), the vanishing gradient point leads to $\lambda_t = -1$ for $t = 1, \ldots, T$. Plugging this constraint on the dual variables back to the (6) and (7) leads us to the following problem:

$$\max_{x \in \{0,1\}^{|\mathcal{V}|}} \sum_{1 \le i < j \le n} x_i x_j \left\{ k(v_i, v_j) - \sum_t \mathbb{I}[(i,j) \notin \mathcal{E}_t] \right\}. \quad (8)$$

---

[1]One could also choose $p = \infty$ which leads to a single global variable $\beta$. However, this choice is not robust against missing edges as a single instance with an empty edge set would render the whole procedure vacuous.

---

**Algorithm 1** $\ell_1$ Soft Clique-ness Measure

---

**Input** A spatio-temporal graph $\mathcal{G}_t(\mathcal{V}, \mathcal{E}_t, k_t)$
Compute the total similarity, $k(i,j) = \sum_t k_t(i,j)$
Compute the measure, $c(i,j) = \sum_t \mathbb{I}[(i,j) \in \mathcal{E}_t]$
Solve $\underset{x}{\text{argmax}} \ \{x^T K x - x^T C x\}$ with $K_{i,j} = k(i,j)$
and $C_{i,j} = c(i,j)$
**Return** $x \in \{0,1\}^{|\mathcal{V}|}$

---

The above problem takes the appealing form of a maximum weighted clique optimization with the readjusted weight functions taking into account the number of missing edges to make group of vertices into a clique. Finding the maximum weighted clique is NP-complete and hard to approximate to a given bound (Pardalos & Xue, 1994). Numerous heuristics have been proposed to obtain local solutions, and eventually the best performing strategy often depends on the application. For a *very* small graph, any integer programming solvers, such as CPLEX, can in principle be used to solve (8). In our case, we deal with graphs of moderate to large size. For this reason, we use the Quadratic Pseudo-Boolean Optimization with Probing (QPBOP) solver[2] of Rother et al. (2007).

**$\ell_2$ Soft Clique-ness Measure** Similarly to the $\ell_1$ case, we find the point where the gradient with respect to $\beta_t$ vanishes leads to the dual variables take the form of $\lambda_t = 2\eta\beta_t$ for (6) and $\lambda_t = -2\beta_t$ for (7). We again use this to eliminate the primal variables $\beta_t$. The optimization problem of (6) is now in the form of[3]:

$$\max_{\lambda \in \mathbb{R}^T_+} \min_{x \in \{0,1\}^{|\mathcal{V}|}} \ - \sum_{1 \leq i < j \leq n} x_i x_j k(v_i, v_j)$$
$$- \frac{1}{4\eta} \sum_t \lambda_t^2 + \sum_t \lambda_t \left\{ \sum_{1 \leq i < j \leq n} x_i x_j \mathbb{I}[(i,j) \notin \mathcal{E}_t] \right\}. \quad (9)$$

The problem (9) is concave quadratic with respect to the dual variables given all the binary indicator variables $x$, and is in the form of standard maximum weighted clique problem with respect to the indicator variables given the dual variables $\lambda$. With these observations, to solve (9), we pursue an alternating approach:

- Step 1: given all the dual variables, solve the following maximum weighted clique problem

---

**Algorithm 2** $\ell_2$ Soft Clique-ness Measure

---

**Input** A spatio-temporal graph $\mathcal{G}_t(\mathcal{V}, \mathcal{E}_t, k_t)$, number of iterations $N$, regularization constant $\eta$
Compute the total similarity, $k(i,j) = \sum_t k_t(i,j)$
**for** $i = 1$ **to** $N$ **do**
  Compute the measure, $c(i,j) = \sum_t \lambda_t \mathbb{I}[(i,j) \notin \mathcal{E}_t]$

  Solve $\underset{x}{\text{argmax}} \ \{x^T K x - x^T C x\}$ with $K_{i,j} = k(i,j)$ and $C_{i,j} = c(i,j)$
  Update $\lambda_t \leftarrow 2\eta \sum_{ij} x_i x_j \mathbb{I}[(i,j) \notin \mathcal{E}_t]$
**end for**
**Return** $x \in \{0,1\}^{|\mathcal{V}|}$

---

$$\min_{x \in \{0,1\}^{|\mathcal{V}|}} \sum_{1 \leq i < j \leq n} x_i x_j \left\{ -k(v_i, v_j) + \sum_t \lambda_t \mathbb{I}[(i,j) \notin \mathcal{E}_t] \right\}.$$

- Step 2: subsequently, given the indicator variables, update dual variables with a closed-form solution, $\lambda_t = 2\eta \sum_{1 \leq i < j \leq n} x_i x_j \mathbb{I}[(i,j) \in \mathcal{E}_t]$.

The above two steps are repeated until certain number of alternating steps is reached. The initialization plays a crucial role for this alternating approach. We set all the dual variables to zeros at the start. Those dual variables that are associated with the time slices where the clique-ness constraints are not violated (there are no missing edges) will stay at zeros (this is a complementary slackness of necessary KKT condition for the optimal points (Boyd & Vandenberghe, 2004)).

## 4. Experiments

We perform two experiments to assess the efficacy of our soft clique-ness measure: first on synthetic data where a collection of noisy snapshots of the same graph is observed, and second on real location-based social network graph where we would like to identify a clique of friends. We discuss both experiments in turn[4].

### 4.1. Synthetic Data

We generate the data in the following ways: a) At time 1, we draw 18 2D data points from a mixture of 3 Gaussian distributions where 7 samples drawn from $\mathcal{N}([0,0], 1.0I)$, 6 samples drawn from $\mathcal{N}([-6,3], 2.0I)$, and 5 points sampled from $\mathcal{N}([8,-3], 2.0I)$, b) For time $t = 2, \ldots, T$, we add to each point of the initial graph, at each time slice, an independent Gaussian noise. This procedure simulates the situation where inconsistent patterns are observed in the several shots

---

[2] http://pub.ist.ac.at/~vnk/software.html
[3] Problem (7) leads to the same form with different constant in front of the term $\sum_t \lambda_t^2$.

[4] The code is available at http://mlg.eng.cam.ac.uk/~nquadrianto/.

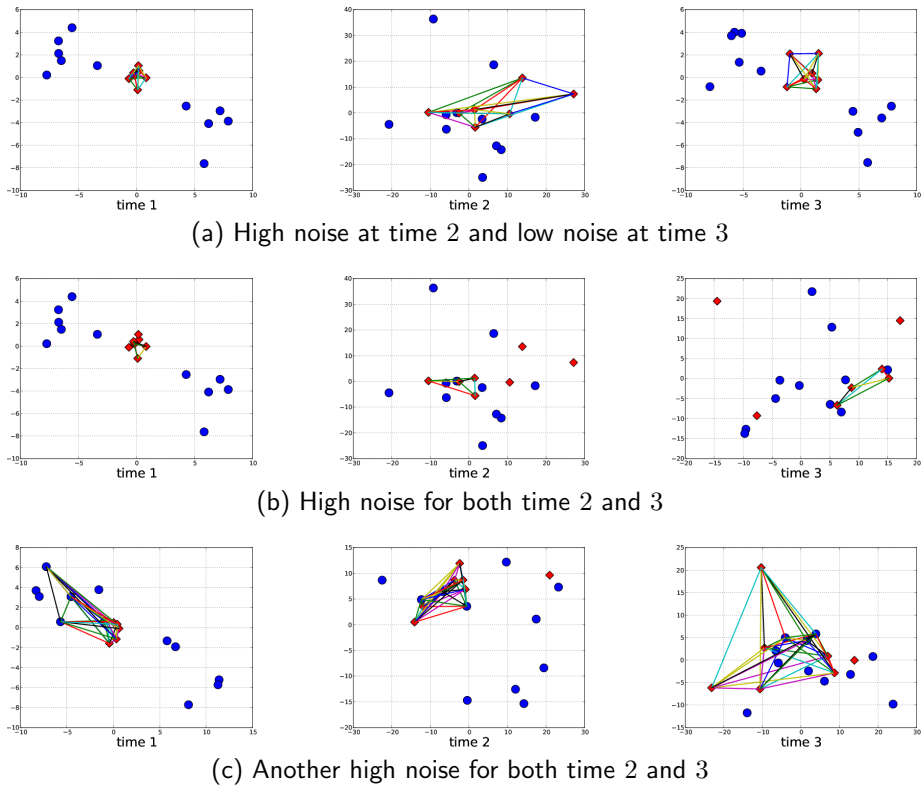

(a) High noise at time 2 and low noise at time 3

(b) High noise for both time 2 and 3

(c) Another high noise for both time 2 and 3

*Figure 2.* Synthetic experiments with data at time 1 are drawn from a mixture of Gaussian distributions with 3 components. At time 2 and 3, the data are corrupted with a random Gaussian noise. Red dots depict the group of vertices that we want to identify as a clique. (a) Under high noise level at time 2 and low noise level at time 3, our method is able to correctly capture the clique. (b) As expected, under high noise levels for time 2 and 3, our approach is only able to recover partial vertices of the clique. (c) Another draw of high noise levels setting.

of the initial graph depending on the amount of noise.

We produce 4 sets of synthetic data with the above procedure. `Syn.Data A` and `Syn.Data B` have 3 time slices where at time 2 a noise of $\mathcal{N}([0,0], 10.0I)$ is added, and at time 3 a noise of $\mathcal{N}([0,0], 0.8I)$ is introduced for A and $\mathcal{N}([0,0], 10.0I)$ is introduced for B. We repeat the data generation process 10 times. Figure 2 shows examples of draws from this process. `Syn.Data C` has 5 time slices and is corrupted by the same noise as A and B at time 2. Subsequently, the noises at time 3, 4, and 5 are $\mathcal{N}([0,0], 2.0I)$, $\mathcal{N}([0,0], 5.0I)$, and $\mathcal{N}([0,0], 0.8I)$, respectively. Lastly, there are 7 time slices at `Syn.Data D`, with the same type of corrupted noises as C for the first four time slices, and the last two time slices have $\mathcal{N}([0,0], 2.5I)$ and $\mathcal{N}([0,0], 0.5I)$ random noises.

We run our $\ell_1$ (refer to Algorithm 1) and $\ell_2$ (refer to Algorithm 2) soft clique-ness measure methods on the generated synthetic data. We use a Gaussian RBF kernel $k(v_i, v_j) = \exp(-\frac{1}{\sigma^2} \|v_i - v_j\|_{\ell_2}^2)$ with the kernel width $\sigma^2$ set to the median distance (Schölkopf, 1997), as the similarity function. Some visualizations of the results of our algorithms are depicted in Figure 2. As a baseline, we use the graph shift algorithm proposed in (Liu & Yan, 2010)[5]. This algorithm finds a mode, i.e. a dense subgraph, on a single graph (Pavan & Pelillo, 2007). We can extend this algorithm to work on samples of graphs by simply finding the mode defined by the temporal average of similarity or affinity matrices. The graph shift returns different local modes depending on different initializations. We use the default setting, which use each vertex as one initialization, and choose the output with the best score. We use the Jaccard Index (Jaccard, 1901), $J(X, \hat{X}) = \frac{|X \cap \hat{X}|}{|X \cup \hat{X}|}$ for a predicted set of fully connected vertices $\hat{X}$ while a true clique is $X$, as an evaluation metric. The empirical results are summarized in Table 1. The soft clique-ness measure is clearly important for handling *inconsistent* patterns inherent in the data. Evidently, our $\ell_1$ and $\ell_2$ soft clique-ness measure methods also produce lower variances in the Jaccard Index metric in comparison to the graph shift algorithm. As expected, our two algorithms will (mostly) recover the same solutions. It is also observed that a collection of snapshots of the same graph could give us more confidence of the clique pattern, for example the mean Jaccard Index in `Syn.Data D` is higher than in `Syn.Data B` .

*Table 1.* Synthetic Experiments Results. Jaccard Index mean ± std over 10 random repeats (the higher the better). `GS`: Graph Shift algorithm (Liu & Yan, 2010); `Soft` $\ell_1$: Our soft $\ell_1$ measure; `Soft` $\ell_2$: Our soft $\ell_2$ measure. For the descriptions of the data, please see texts.

| Data | GS | Soft $\ell_1$ | Soft $\ell_2$ |
|------|----|----|----|
| `Syn.Data A` | 0.82±0.28 | 0.92±0.12 | 0.89±0.14 |
| `Syn.Data B` | 0.58±0.31 | 0.62±0.16 | 0.64±0.17 |
| `Syn.Data C` | 0.79±0.28 | 0.88±0.18 | 0.87±0.16 |
| `Syn.Data D` | 0.85±0.26 | 0.93±0.12 | 0.89±0.15 |

### 4.2. Real Social Network Data

In this experiment, we are interested to apply our clique-ness measure to the samples of real network graphs for identifying a soft-clique of friends.

**Data** We use a Brightkite location-based social network graph[6]. The network contains data of users' "check-ins" where users shared their locations by using text messaging or other mobile applications. The numbers of check-ins varies widely among users, thus we filter the data and only use data from persons with above average number of check-ins. This leaves us with 4,429 persons or nodes and 9,805,806 connections.

**Set Kernels** Exploiting studies in Cho et al. (2011) that mobility patterns of humans are mostly periodic (moving back and forth between homes and workplaces) and the remaining patterns could be explained by social relationships, we define different *after*-hours in a day as samples of the graphs. Specifically, we uniformly divide the hours between 17 : 00 and midnight into 7 time slices. Each person or node now is represented as a vector of location and date of checkin. To define kernels between persons in this (location-date)-based representation, we are faced with the challenge that persons are represented by different numbers of locations at different dates depending on how (ir)regular their check-ins behaviors. We pursue a solution of representing a person with a set or a bag of vectors. In this paper, we use the following set kernel between $x = \{x_1, \ldots, x_n\}$ and $y = \{y_1, \ldots, y_m\}$: $k_{\text{set}} = \sum_{i=1}^{n} \sum_{j=1}^{m} k_{\text{base}}(x_i, y_j)$, where $x_i$ and $y_i$ are the persons (location-date) descriptors. A Gaussian RBF is used as the base kernel.

**Diagonal Dominance** We observe that our set kernel matrices at each time slice are diagonally dominant, i.e. a person tends to be *much* more similar to himself or herself than to others. Sub-polynomial kernels (Schölkopf et al., 2003) are designed to specifically address such diagonal dominance issue. This sub-polynomial kernel is generated as follows: first each element of the original kernel matrix is raised to the power of $p$ where $0 < p < 1$. This procedure reduces the dynamic range of the original kernel as elements that are less than one will be increased, whereas elements that are greater than one will be decreased. Then, the rows of the modified kernel is normalized to a unit length (let $\hat{K}$ denote this transformed kernel matrix), and lastly, our sub-polynomial kernel matrix is $\hat{K}\hat{K}^\top$. The last step is needed to make the modified kernel matrix to be a positive definite matrix.

**Results** Our $\ell_1$ method run for about an hour and produce a soft-clique of friend of the size 1754. We observe that our identified clique explains 23% of the friendship network that was collected based on the online public API. The graph shift algorithm did not finish in a week time. We thus randomly select 1754 nodes to form a clique and find that the random clique explains only 14% of the online friendship network.

## 5. Discussion and Conclusion

We have introduced the concept of *persistent soft-cliques*, i.e., subgraphs that are almost fully or at least densely connected, and that persists through all or most instances of the graph. This concept is particularly useful to characterize key patterns more confidently when multiple instances of a graph are available for analysis. We presented a soft clique-ness measure, that counts and penalizes the number of edges missing to make the selected subgraphs into a clique, to handle inconsistent patterns inherent in noisy graph instances. When this measure is used in an optimization framework to find a maximum weighted clique, we end up at a two-person game or a slack formulation. We showed using a Lagrangian method that the two formulations lead to the same solution.

Our experiments on randomly generated inconsistent patterns in several shots of the initial graph confirmed that a collection of, though noisy, snapshots of the same graph can give us more confidence of the clique pattern, and our clique-ness measure is important for such situation. We also provided experiments on a prototypical application of our method: identifying groups of friends or families from a location-based social network graph where the sampling of graphs correspond to an *after*-hour temporal dimension. Our clique-ness measure helped to recover more than 20% of the online friendship network.

Despite encouraging results in our experiments, we have clearly only touched the surface of possibilities to be explored. Particularly, we are interested to explore the case where vertex correspondences between time slices are a priori unknown. In such situation, we need to *jointly* infer the key patterns and the correspondences, specifically a mutual information like dependency measure can be used for the latter. Furthermore, we intend to introduce a pairwise coupling over time slices via a *Markov model*. This model will allow us to distinguish the case where edges disappear for a while and then come back to the case where edges randomly flip on and off.

### Acknowledgments

The authors would like to thank Ahmed Jawad, Marcello Pelillo, Zoubin Ghahramani, Adrian Ion, Tiberio Caetano, and Viktoriia Sharmanska for discussions. The first author is supported by a Newton International Fellowship.